\pdfoutput=1

\documentclass[11pt]{article}

\usepackage[]{acl}

\usepackage{times}
\usepackage{latexsym}

\usepackage[T1]{fontenc}

\usepackage[utf8]{inputenc}

\usepackage{microtype}

\usepackage{inconsolata}

\usepackage{amsmath}
\usepackage{graphicx}
\usepackage{subcaption}
\usepackage[normalem]{ulem}
\useunder{\uline}{\ul}{}
\usepackage{threeparttable}
\usepackage{lipsum}
\usepackage{enumerate}
\usepackage{booktabs}
\usepackage{multirow}
\usepackage{amssymb}
\usepackage{bookmark}
\usepackage{color}  
\definecolor{shadecolor}{rgb}{0.92,0.92,0.92}  
\usepackage{framed} 
\usepackage{cases}
\usepackage{tcolorbox}
\usepackage{xcolor}
\newtcolorbox{myshaded}{
  colframe=black,
  colback=gray!10,
  boxrule=0.5pt,
  left=5pt, 
  right=5pt, 
  top=5pt, 
  bottom=5pt, 
}

\title{RepEval: Effective Text Evaluation with LLM Representation}

\author{Shuqian Sheng\textsuperscript{1}, Yi Xu\textsuperscript{1}, Tianhang Zhang\textsuperscript{1}, Zanwei Shen\textsuperscript{1}, Luoyi Fu\textsuperscript{1\thanks{* Luoyi Fu is the corresponding author.}},\\\ \textbf{ Jiaxin Ding\textsuperscript{1}, Lei Zhou\textsuperscript{1}, Xiaoying Gan\textsuperscript{1}, Xinbing Wang\textsuperscript{1}, Chenghu Zhou\textsuperscript{2}}\\
\textsuperscript{1}Shanghai Jiao Tong University, Shanghai, China\\
\textsuperscript{2}IGSNRR, Chinese Academy of Sciences, Beijing, China\\
  {\tt \{susisheng, yixu98, zhangtianhang, yiluofu\}@sjtu.edu.cn} \\  }

\begin{document}
\maketitle
\begin{abstract}
The era of Large Language Models (LLMs) raises new demands for automatic evaluation metrics, which should be adaptable to various application scenarios while maintaining low cost and effectiveness. Traditional metrics for automatic text evaluation are often tailored to specific scenarios, while LLM-based evaluation metrics are costly, requiring fine-tuning or rely heavily on the generation capabilities of LLMs. 
Besides, previous LLM-based metrics ignore the fact that, within the space of LLM representations, there exist direction vectors that indicate the estimation of text quality. 
To this end, we introduce RepEval, a metric that leverages the projection of LLM representations for evaluation. Through simple prompt modifications, RepEval can easily transition to various tasks, requiring only minimal sample pairs for direction vector construction. Results on fourteen datasets across two evaluation tasks demonstrate the high effectiveness of our method, which exhibits a higher correlation with human judgments than previous methods, even in complex evaluation scenarios involving pair-wise selection under nuanced aspects. Our work underscores the richness of information regarding text quality embedded within LLM representations, offering insights for the development of new metrics.
\footnote{The project is publicly available for research purpose 
    https://github.com/susisheng/RepEval}
\end{abstract}

\section{Introduction}
\label{sec:introduction}
Text evaluation is widely applied in the era of LLM, such as detecting harmful responses~\cite{sun2023safety, kim2024prometheus}, identifying high-quality data for model training~\cite{llama3, cai2024internlm2} and constructing preference data for model alignment~\cite{nemo, bai2022constitutional, lee2023rlaif}.
Such requirements pose significant challenges to automatic text evaluation metrics, as metrics must be adaptive to diverse evaluation tasks and achieve high-quality assessment while maintaining a low cost.
However, traditional automatic evaluation metrics, such as BLEU~\cite{Papineni-2002-bleu} and COMET~\cite{rei-etal-2020-comet}, 
are usually designed for specific tasks or criteria, making them difficult to transfer to new application scenarios. Also, their requirement for references and other inputs makes them infeasible in various evaluation contexts. 
LLM-based metrics offer a possible solution~\cite{gao2023humanlikegpt, chiang_can_llm_2023}, but such metrics may also encounter certain limitations. On the one hand, they rely heavily on the generation ability of LLM to adhere to predefined formats, which typically require more model parameters or fine-tuning, resulting in higher costs for inference and deployment. On the other hand, their assessment is frequently unsatisfactory, which does not align well with human judgments and exhibits unstable performance~\cite{shen-etal-2023-llm-summ-eval}. 

Fortunately, though language models may struggle to generate appropriate responses, their representations contain rich information related to correct answers, which could be extracted with neural network or other models~\cite{zou2023representation}.
Imagine, when people are assessing a piece of text, they may have a clear sense of its quality yet struggle to quantify their impressions with a precise score.
This implies that during evaluation, we can reduce the reliance on the generation capabilities of LLMs and instead focus on the meaningful information contained in their representations. By doing so, we can utilize models with fewer parameters, thereby avoiding excessive computational resource consumption while achieving better performance. The remaining questions are: Do representations of LLM really encapsulate information relevant to text quality? How can we effectively \textbf{extract and apply} this information to evaluation tasks?

In this study, we introduce \textbf{RepEval}, a metric utilizing the projection of LLM representation for custom evaluation. We explored the performance of RepEval in two scenarios: absolute evaluation and pair-wise evaluation. In \textbf{absolute evaluation}, which requires evaluation metrics to output scores as assessment, our intuition is that representations of high-quality and low-quality text exhibit distinct distributions. 
We validate that, in vector space, their projection in a specific direction characterizes the degree of variation in textual properties. 
In \textbf{pair-wise evaluation}, metrics need to select the better one out of the two inputs.
To solve this problem, we construct a projection vector that measures the probability of whether the preceding sentence is better than the latter.

For absolute evaluation, experiments on three criteria with ten datasets show that our method has better correlations with human judgments than previous metrics, which is flexible and easy to extend to other applications. As to pair-wise evaluation, experiments on four tasks with custom criteria demonstrate that our method remains highly feasible in complex application scenarios, achieving excellent classification accuracy. 
Through visualization, we further demonstrate that a well-designed prompt can transfer the representation to different positions within the semantic space, thus facilitating evaluations based on diverse criteria.
We also demonstrate that using PCA can produce nearly optimal projection vectors, and we explore the optimization strategy of RepEval, offering a reasonable scheme for representation creation.

In summary, the key contributions of this work are:

\begin{itemize}
    \item We introduce the evaluation metric RepEval, surpassing previous metrics on nearly all tasks, even outperforming GPT-4 with much fewer model parameters.
    \item RepEval can easily adapt to new evaluation scenarios, requiring only a few samples for training, and obviating the need for extensive human annotations and LLM fine-tuning.
    \item RepEval offers insights for the introduction of new metrics, demonstrating that LLM representations contain decisive information about text quality inherently.
\end{itemize}

\section{Related Work}
Automatic evaluation metrics can be categorized into three types, reference-based, reference-free, and LLM-based metrics.
\subsection{Reference-based metrics}
Reference-based metrics measure the similarity between the hypothesis and one or multiple references, and a hypothesis more similar to the reference is considered to be better~\cite{gehrmann2023repairing}. These metrics can be further classified into two types: n-gram-based and embedding-based. Popular n-gram-based metrics include BLEU~\cite{Papineni-2002-bleu}, ROUGE~\cite{lin2004rouge} and METEOR~\cite{banerjee-lavie-2005-meteor}. Embedding-based metrics include BERTScore~\cite{zhangbertscore} and MoverScore~\cite{zhao-etal-2019-moverscore}. However, the requirement of human-written references limits their applications, as the creation of references is always a serious problem.
\subsection{Reference-free metrics}
Reference-free metrics instead require the source to generate the hypothesis in the Natural Language Generation(NLG) process. Their advantage lies in the independence of human-written references, which costs expensive manual preparation. Polular reference-free metrics include BARTScore~\cite{NEURIPS2021_bartscore}, UniEval~\cite{zhong-etal-2022-unieval} and GPTScore~\cite{fu2023gptscore}. Compared to reference-based metrics, they often exhibit better performance and adaptability~\cite{sheng2024reference}. However, these metrics are mostly designed for specific application scenarios and criteria, making it challenging to effectively apply them to new tasks.

\subsection{LLM-based metrics}
In recent years, there is a new trend to utilize LLM in text evaluation. Relying on the powerful capabilities of LLM, these studies use few-shot or zero-shot methods to directly generate the assessment results~\cite{gao2023humanlikegpt, chiang_can_llm_2023}. To enhance model performance, some studies have trained models specifically for evaluation through fine-tuning~\cite{kim2024prometheus}. However, LLMs with better generation capability usually contain more parameters, which is costly for evaluation, while the outputs are often unsatisfactory~\cite{shen-etal-2023-llm-summ-eval}. The method of fine-tuning is also time-consuming and expensive as well.

\begin{figure*}
    \centering
    \includegraphics[width=0.98\textwidth]{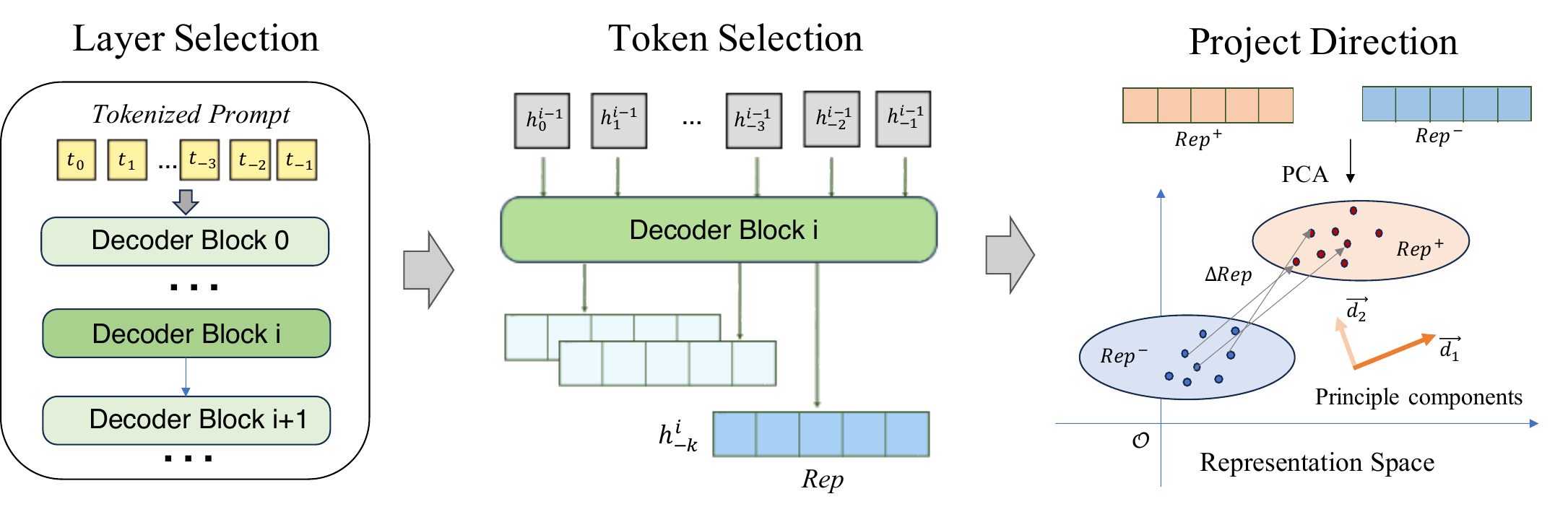}
    \caption{Pipeline of collecting representations with decoder-only LLM and constructing project direction.}
    \label{fig:rep_pipeline}
\end{figure*}

\section{Preliminary}

\subsection{Standard Evaluation}
\label{sec:standard_evaluation}
A standard NLG process receives a source text $src$ as input and outputs a text $hyp$ based on certain requirements, which can be seen as a generation function. In the same scenarios, an answer written by human experts can be viewed as a reference $ref$. 

A common evaluation scenario is \textbf{absolute evaluation}, where an automatic metric function $f$ is applied to evaluate a single $hyp$ based on the specific criterion and output the evaluation result in the form of a score. This process can be described as Equation~\ref{eq:metric_score_function}. We should note that $src$ and $ref$ are not necessary for all metrics. Also, for some metrics, the evaluation scores are irrelevant to the criterion.

\begin{equation}
\label{eq:metric_score_function}
    score = f(criterion, hyp, src, ref)
\end{equation}

Another scenario is \textbf{pair-wise evaluation}. Each time in the evaluation, a pair of $hyp$ is provided, and metrics are required to choose the better one from two $hyp$s based on specific criteria. Datasets in this scenario are all collected from complicated tasks, which have custom evaluation criteria for different samples. This scenario requires the model to clearly understand the evaluation criteria and accurately discern the quality difference between $hyp$ pairs.

\subsection{Meta-Evaluation}
Human judgment is still the gold-standard approach to text evaluation~\cite{NEURIPS2021_bartscore}, which is also the basis of meta-evaluation methods used in this study.

In absolute evaluation tasks, the effectiveness of the metric is measured by the correlation between its scores and human judgments. The calculation is shown in~\ref{eq:correlation}.
\begin{equation}
\label{eq:correlation}
    \begin{split}
        correlation = \rho([s_1, s_2, \ldots, s_N], \\
        [h_1, h_2, \ldots, h_N])
    \end{split}
\end{equation} where $s_i$ is the metric score of the i-th sample in a certain dataset, $h_i$ is the relative human judgment, and $\rho$ is the correlation function. In this study, we use Spearman Correlation~\cite{Spearman1987ThePA}. 

In pair-wise evaluation scenarios, we use the accuracy of detecting better $hyp$ as the meta-evaluation method, as shown in Equation~\ref{eq:accuracy}
\begin{equation}
    \label{eq:accuracy}
    accuracy = \frac{1}{N} \sum_{i=1}^{N} \mathbb{I}(\hat{y}_i = y_i)
\end{equation}
Where $N$ is the number of sample pairs, $\hat{y}_i$ is the predicted index of better $hyp$, and $y_i$ is the ground truth label.

\subsection{Representations of LLM}
\label{sec:llm_representation}
In this study, representation refers to the hidden states of LLM with specific input texts. LLMs utilized in this study are in decoder-only architecture, typically comprising $n$ decoder layers and a language modeling head with the hidden size of $d$. As shown in Figure~\ref{fig:rep_pipeline}, specifically, given a text input with $s$ tokens, denote the output of the $i$th layer as $h^i$, where $i \in [0, n-1]$, and $h^i \in \mathbb{R}^{s \times d}$. We further denote the hidden states of $k$th token on layer $i$ as $h_k^i$. 

Suppose we choose the last $k$th token on the $i$th layer as the representation $rep$, we then have $rep = h_k^i$. 

\begin{figure*}[!h]
    \centering
    \includegraphics[width=0.98\textwidth]{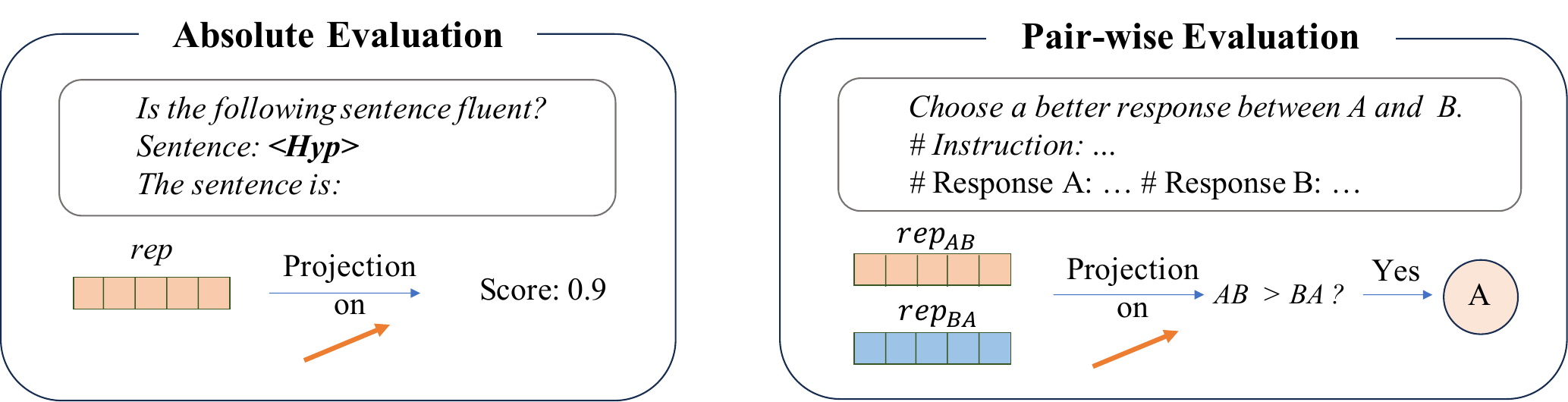}
    \caption{Evaluation process of absolute evaluation and pair-wise evaluation.}
    \label{fig:rep_score}
\end{figure*}

\section{Methodology}
\label{sec:methodology}

\subsection{Collecting Representation}
\label{sec:collect_rep}
Though RepEval does not rely on the generation ability of LLM, a good prompt helps integrate representations with more information related to the evaluation tasks. As defined in Section~\ref{sec:standard_evaluation}, to collect the representation $rep$, we can simply apply $hyp$ as input. However, this is agnostic to the evaluation scenarios, and constructing task-related prompt templates helps improve the performance. 

For absolute evaluation, we adopt three general criteria: fluency, consistency and coherence. In this scenario, the metric score represents how likely the $hyp$ is a qualified text. We design and utilize the following prompt template.

\begin{myshaded}  
\textit{Is the following Hyp <criterion\_description>? }

\textit{Hyp: <hyp>}

\color[RGB]{117, 113, 113} {
\textit{Src: <src> }
}

\color[RGB]{0, 0,0}
\textit{The sentence is}
\end{myshaded}

Here, ``<hyp>'' is filled by $hyp$ to be evaluated, ``<src>'' is optional and only used in consistency evaluation, while ``<criterion\_description>'' is different for each criterion. Please refer to the Appendix~\ref{sec:repeval_prompt} for more information.
We also add a control group without the prompt template, using only $hyp$ as inputs.

For pair-wise tasks, we need to compare the quality of two different $hyp_A$ and $hyp_B$. Datasets related to the pair-wise evaluation are collected from complicated tasks, adopting different score criteria for each sample, such as harmlessness, honesty, etc. 
Follows~\citet{kim2024prometheus},
here, ``<instruction>'' is the description of the task description, ``<response 1>'' and ``<response 2>'' could be filled by
$hyp_A$ and $hyp_B$, and ``<score criterion>'' is the evaluation requirement. More details could be found in the Appendix~\ref{sec:repeval_prompt}
\begin{myshaded}  
\textit{Instruction: <instruction>}

\textit{Response A: <response 1>}

\textit{Response B: <response 2>}

\textit{Score Rubric: <score criterion>}

\textit{Ans:}
\end{myshaded}

By exchanging the position of $hyp_A$ and $hyp_B$ in the prompt, we can obtain two $rep$s, marked as $rep_{AB}$ and $rep_{BA}$. These $rep$s contain information about the following question: How likely is the previous sentence better than the latter? 
We will explain how to utilize this information in subsequent sections.

\begin{table*}[t]
\centering
\small
\tabcolsep 0.041in
\begin{tabular}{cc|ccc|c|ccc|ccc|c}
\toprule
                             &           & \multicolumn{4}{|c|}{RepEval}                          & \multicolumn{6}{c}{Baselines}                                          \\
\cmidrule{3-13}
                             &           & \multicolumn{3}{|c|}{Prompt}                       & Hyp-only    & \multicolumn{3}{c|}{LLM}  & \multicolumn{3}{c|}{Ref-free}                     & \multicolumn{1}{c}{Ref-based}                \\
\cmidrule{3-13}{\tiny}
                             &    & PCA(20)      & PCA(5)       & SVM            & PCA(20)   &  GPT-4 & GPT-3.5 &  Mistral-7b & GPTS       & BARTS      & UniE           & BertS   \\
\midrule
\multirow{8}{*}{\textbf{FLU}}     & BAGEL     & \textbf{0.330} & 0.236          & \textbf{0.358} & 0.060    &0.325    & 0.222 & 0.156 & 0.152          & 0.241          & 0.309            & 0.247  \\
 & Newsroom  & 0.548          & \textbf{0.565} & 0.515          & 0.478   &0.297     & 0.218 & 0.411 & 0.565          & \textbf{0.596} & 0.443            & 0.182  \\
 & SFHOT     & \textbf{0.351} & 0.345          & \textbf{0.368} & 0.108   &0.305       & 0.178 & 0.238 & 0.135          & 0.164          & 0.312            & 0.164  \\
 & SFRES     & \textbf{0.377} & 0.370          & \textbf{0.391} & 0.021   &0.352       & 0.289 & 0.272 & 0.229          & 0.226          & 0.332            & 0.183  \\
 & SummEval  & \textbf{0.447} & 0.424          & 0.419          & 0.324   &0.245       & 0.120 & 0.285 & 0.288          & 0.285          & \textbf{0.451}   & 0.194  \\
 & USR-P     & 0.360          & \textbf{0.404} & 0.363 & 0.306   &\textbf{0.391}       & 0.310 & 0.288 & -0.030         & 0.034          & 0.239            & 0.322  \\
 & USR-T     & 0.329          & \textbf{0.368} & 0.336          & \textbf{0.402} &0.324 & 0.203 & 0.309 & 0.087          & 0.027          & 0.302            & 0.292  \\
 & WebNLG    & \textbf{0.587} & 0.534          & \textbf{0.633} & 0.268   &0.503       & 0.409 & 0.401 & 0.072          & 0.330          & 0.521            & 0.499  \\

\midrule
\multirow{3}{*}{\textbf{CON}} & QAGS-C  & 0.541          & 0.561          & 0.453          & NA      &0.505       & 0.295 & 0.380 & 0.583          & \textbf{0.680} & \textbf{0.618}   & 0.507  \\
 & QAGS-X & 0.497          & \textbf{0.550} & \textbf{0.524} & NA    &0.457         & 0.315 & 0.185 & 0.081          & 0.159          & 0.387           & -0.057 \\
 & SummEval  & 0.426 & 0.421          & 0.342          & NA    &\textbf{0.436 }        & 0.269 & 0.210 & 0.355          & 0.334          & \textbf{0.435}   & 0.200  \\

\midrule
\multirow{2}{*}{\textbf{COH}}   & Newsroom  & 0.444          & 0.392          & 0.273          & 0.373    &0.274      & 0.207 & 0.421 & \textbf{0.595} & \textbf{0.623} & 0.458   & 0.221  \\
 & SummEval  & \textbf{0.534} & 0.516          & 0.418          & 0.263   &0.347       & 0.247 & 0.262 & 0.412          & 0.408          & \textbf{0.592}   & 0.333 \\
\bottomrule
\end{tabular}

\caption{\textbf{Absolute Evaluation Results.} Each row represents the \textbf{Spearman's correlations} of a metric with human judgments on absolute evaluation datasets. The \textbf{bold} scores represent the top two highest correlation results for each task on each criterion. Coherence, consistency, and fluency are written in abbreviations COH, CON, and FLU respectively. PCA(n) represents $n$ samples are used in training. Hyp-only can not be used for consistency evaluation.
}
\label{tab: spearman_all}
\end{table*}

\subsection{Project Direction}
\label{sec:project_direction}
In the previous steps, we converted both evaluation tasks into \textbf{binary classification} problems by constructing proper prompts and obtained the relevant $rep$s. Next, we need to figure out a specific projection direction $\Vec{d}$, where the projection of $rep$ on $\Vec{d}$ represents the probability of the answer is ``Yes''. 

We utilize Principal Component Analysis (PCA) to accomplish this task.
In absolute evaluation, assume we have $K$ high-quality texts, i.e. they receive high scores from human evaluators, and we denote their representations as $rep^+$. Similarly, we collect $K$ low-quality texts and their representations, denoted as $rep^-$. For each pair of $(rep^+, rep^-)$, their difference is given by $\Delta rep = rep^+ - rep^-$ or $\Delta rep = rep^- - rep^+$. 
In pair-wise evaluation, consider $K$ pairs of texts, where one sentence (A) is better than the other (B). According to the process described in section~\ref{sec:collect_rep}, since A is better than B, we denote the representation of $rep_{AB}$ as $rep^+$ and $rep_{BA}$ as $rep^-$. Here, $\Delta rep$ indicates the probability that 'A is better than B'."

As shown in Figure~\ref{fig:rep_pipeline}, $\Delta rep$s represents the change in the likelihood of the answer being ``Yes'' instead of ``No'' in each sample, while their principal components should capture the overall variations. Therefore, with $\Delta rep$ as inputs, assuming that we collect k main component vectors with PCA, as well as their importance score. Mark the $i$th vector and its importance as $\Vec{d_i}$ and $w_i$. 
we can obtain the final $\Vec{d}$ following Equation~\ref{eq:pca_vector}:

\begin{equation}
    \Vec{d} = \sum^k_{i=1} w_i\Vec{d_i}
    \label{eq:pca_vector}
\end{equation}

\subsubsection{Collect Evaluation Results}
\label{sec:collect_score}
As shown in Figure~\ref{fig:rep_score}, we obtain the evaluation score following equation~\ref{eq:rep_score_abs} in absolute evaluation.
\begin{equation}
\label{eq:rep_score_abs}
    score = {rep}^T\Vec{d}
\end{equation} where ${rep}$ is the representation of the $hyp$, $\Vec{d}$ is the project direction vector, marked the probability of $hyp$ been a qualified text.

In pair-wise evaluation, by switching the position of $hyp_A$ and $hyp_B$, we obtain two $rep$s, noted as $rep_{AB}$ and $rep_{BA}$, respectively. Then the prediction result is:
\begin{equation}
    prediction = \begin{cases}
    A, & rep_{AB}^T\Vec{d} > rep_{BA}^T\Vec{d} \\
    B, & else
    \end{cases}
\end{equation}
where $\Vec{d}$ is the project direction, marking the probability of the first $hyp$ better than the latter one.

\subsection{Selection of layer and token}
As shown in Figure~\ref{fig:rep_pipeline}, when constructing representations, there are many layers and tokens to choose from, and the optimal layer may depend on specific tasks and input. As we only utilize decoder-only LLM, which predicts the next token from left to right, the $rep$s of the last few tokens contain the semantic information of the entire preceding text and are selected for application.

After projection vectors are collected, we test the performance of different tokens combined with different layers, and select the target token and layer with the best performance, i.e. with the highest human correlations or pair-wise accuracy, on the validation set, and apply it to the test set.

\section{Experiments}
\label{sec:experiment}

\begin{table*}[t!]
\centering
\small
\tabcolsep 0.065in
{\begin{tabular}{lccccccccc}
    \toprule

    \multicolumn{1}{c}{\multirow{2}{*}{\textbf{Evaluator LM}}}& \multicolumn{5}{c}{\textsc{HHH Alignment}} & \multicolumn{1}{c}{\textsc{MT Bench}} & \multicolumn{1}{c}{\textsc{Auto-J }} & Preference Bench\\ 
    \cmidrule(lr){2-6} \cmidrule(lr){7-7} \cmidrule(lr){8-8} \cmidrule(lr){9-9} & Help. & Harm. & Hon. & Other & Total Avg. &  w/o TIE &  w/o TIE & Instance-wise\\ 
    \midrule
\textsc{Llama2-Chat 7B}&55.93&62.07&49.18&62.79&57.01&50.39&45.73 & 58.60\\
\textsc{Llama2-Chat 13B}&71.19&77.59&60.66&62.79&68.33&49.61&43.28 & 63.00\\
\textsc{Llama2-Chat 70B}&62.71&81.03&65.57&65.12&68.78&60.88&50.64 & 64.70\\
\textsc{Mistral-Instruct-7B}&59.32&68.97&63.93&81.40&67.42&63.82&60.94 & 79.40\\
\textsc{Mixtral-Instruct-8x7B}&83.05&87.93&67.21&69.77&77.38&{71.42}&73.50 & 84.00\\
\textsc{Pair RM (0.4B)}&{84.75}&84.48&{80.33}&{90.70}&{84.62}&59.00&59.05 & 81.80\\
\textsc{Ultra RM (13B)}&{86.44}&79.31&{81.97}&{88.37}&83.71&56.00&59.85 & 86.97\\
\textsc{Auto-J (13B)}&77.97&79.31&70.49&74.42&75.57 & 69.12 & {76.64} & 81.35\\
\textsc{Prometheus-2-7B}&76.27&{87.93}&73.77&76.74&78.73&67.25&73.80 & {\textbf{92.45}}\\
\textsc{Prometheus-2-8x7B}&84.75&96.55&{81.97}&76.74&85.52&{71.96}&{79.98} & {\textbf{90.65}}\\
\midrule
\textsc{RepEval(pair5)}&89.83&96.55&\textbf{95.08}&\textbf{100.00}&\textbf{95.00} & \textbf{79.90}&73.11 &87.20\\
\textsc{RepEval(pair20)}&\textbf{93.22}&\textbf{100.00}&\textbf{98.36}&\textbf{100.00}&\textbf{97.74}&\textbf{80.39}&74.98&87.90\\
\midrule
\textsc{GPT-3.5-Turbo-0613}&77.97&81.03&77.05&67.44&76.47&69.41&72.13 & 75.05\\
\textsc{GPT-4-1106-Preview}&89.83&96.55&91.80&83.72&90.95&\textbf{79.90}&\textbf{83.12} & 85.50\\
\textsc{Claude-3-Opus}&\textbf{91.53}&\textbf{100.00}&91.80& 95.35&94.57&77.65&\textbf{82.92}&89.85\\
\bottomrule    
\end{tabular}}
\caption{\textbf{Pair-wise Evaluation Results.} Each row represents the \textbf{accuracy} 
(\%) of a metric on selecting better $hyp$ based on specific criteria. The \textbf{bold} scores represent the top two highest accuracy results for each evaluation task. PCA(n) represents $n$ samples are used in training.}
\label{tab:pair-wise-eval}
\end{table*}

\subsection{Datasets}
For absolute evaluation, we focus on three evaluation criteria: fluency, consistency and coherence, which are widely applied in NLG tasks. We utilize datasets from four tasks: Asset~\cite{alva-manchego-etal-2020-asset} for simplification, SummEval~\cite{fabbri-etal-2021-summeval} and Newsroom~\cite{grusky2018newsroom} for summarization, WebNLG~\cite{shimorina2019webnlg}, SFRES, and SFHOT~\cite{wen-etal-2015-semantically} for data-to-text, and USR-Persona and USR-Topic for dialogue~\cite{mehri-eskenazi-2020-usr}.

For pair-wise evaluation, according to~\citet{kim2024prometheus}, we utilize datasets HHH Alignment~\cite{askell2021hhhalign}, MT Bench Human Judgment, Auto-J Eval~\cite{li2023autoj}, and Preference Bench~\cite{kim2024prometheus}. All samples in these datasets contain a pair of $hyp$s, instructions to generate the $hyp$, human judgments and relevant criteria.  For both scenarios, all texts in datasets are written in English. 
\subsection{Baselines}
For absolute evaluation, we utilize three reference-based metrics BERTScore~\cite{zhangbertscore}, along with three reference-free metrics: GPTScore~\cite{fu2023gptscore}, BARTScore~\cite{NEURIPS2021_bartscore}, and UniEval~\cite{zhong-etal-2022-unieval}. Additionally, we employ the Mistral-7b model\footnote{https://huggingface.co/mistralai/Mistral-7B-Instruct-v0.2} and the ChatGPT API (gpt-3.5-turbo and gpt-4)
provided by OpenAI to establish baselines by prompting LLMs for evaluation, following the approach of~\citet{shen-etal-2023-llm-summ-eval}. Baseline For pair-wise evaluation are direct generation results of different LLMs, referencing from~\cite{kim2024prometheus}. Please refer to Appendix~\ref{app:exp} for more details about datasets and metrics.

\subsection{Training Dataset}
We utilize Asset and GCDC for absolute evaluation. Asset belongs to the simplification task, while GCDC is a real-world text dataset specifically created for coherence evaluation, both unrelated to other datasets in this work. 
Please refer to the Appendix for how we select positive samples and negative samples to construct $rep^+$ and $rep^-$.

Since the criteria and application scenarios of pair-wise datasets differ greatly from each other, they can be regarded as unrelated external data. Therefore, for the evaluation of MT Bench Human Judgment, Auto-J Eval, and Preference Bench, we utilize HHH Alignment to construct a training set. For the evaluation of HHH Alignment, we utilize the MT Bench as training data. 
\begin{figure*}[t]
  \centering
  
  \includegraphics[width=0.98\textwidth]{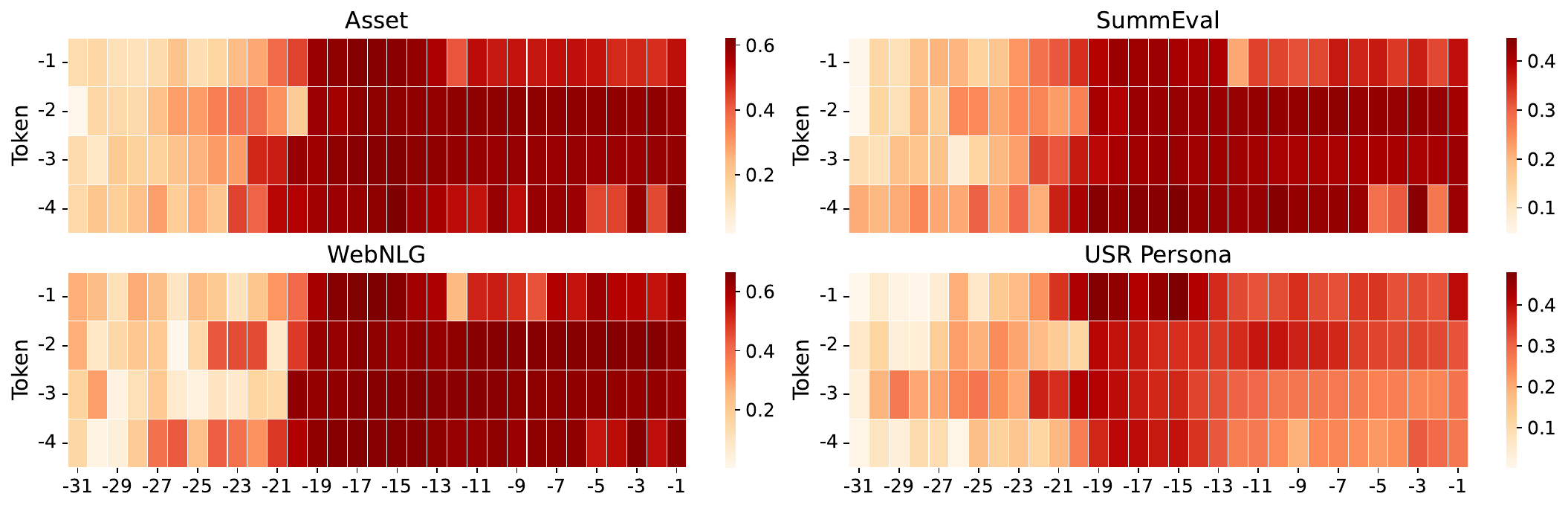}
  \caption{Correlation results for the absolute evaluation of fluency using RepEval with different token and position selections. Layer and token counts are in reverse order, measuring the distance from the output. For instance, layer=-1 represents the last layer closest to the output.}
\label{fig:heatmap}
\end{figure*}
\subsection{Absolute Evaluation}
Following the description in previous sections, the correlations between human judgments and scores generated by each metric are presented in Table~\ref{tab: spearman_all}.

We observe that RepEval outperforms existing metrics on almost all datasets, even surpassing the performance of GPT-4. With just five text pairs, the PCA method surpasses all baseline metrics on half of the datasets, and with 20 pairs, it achieves a top-two performance on seven datasets, similar to the results obtained by SVM. Considering that the training of SVM requires much more samples to achieve similarly good results, PCA significantly reduces the manual cost of constructing samples while maintaining relatively good performance. The Hyp-only experiment's outcome indicates that even without the addition of a prompt template, the embeddings in LLM contain information related to evaluation criteria such as fluency and coherence. Another notable point is that RepEval's performance is evidently better than directly prompting Mistral-7b for evaluation, indicating that even when LLM struggles to generate a satisfying response, their representations can still convey valuable information.

In summary, the projection of $rep$s can efficiently extract information related to the text quality on the desired evaluation criterion of $hyp$ with a few samples. 
Therefore, in most cases, there's no need to employ more complex models like SVM. Additionally, RepEval only requires $hyp$ as input, whereas traditional metrics depend on $src$ or $ref$. 
Compared with directly prompting LLMs like GPT-4, it exhibits better performance while maintaining a relatively low computational and time cost.

\subsection{Pair-wise Evaluation}
The accuracy of each method in pair-wise evaluation is presented in Table~\ref{tab:pair-wise-eval}.
We can observe that despite the varying generation tasks and evaluation criteria for each sample, RepEval still achieves high accuracy in selecting the better $hyp$.
Compared to the generation results of vanilla Mistral-7b, the improvement of RepEval in pair-wise evaluation further validates that, failing to generate a good response does not mean that LLM doesn't know the answers, as $rep$s already contain clear directions pointing towards the correct classification within the semantic space. Moreover, RepEval only adopts general LLM that has not been fine-tuned on evaluation tasks. Compared to PROMETHEUS, which is a text evaluation LLM fine-tuned with millions of data, our method saves the expensive cost of training, while maintaining relatively good or better performance. At the same time, by using only a 7b model, RepEval is still comparable to or even surpasses LLMs like GPT-4.

The above experimental results demonstrate that when there is no need to explain the judgment results, RepEval is highly competitive and can accurately make pair-wise selections. By only using a general LLM for inference, RepEval eliminates the high costs associated with pre-training. Additionally, since the optimal layers often reside in the middle layers, it reduces both inference time and computational costs by not requiring the inference of all parameters.

\begin{figure}[!h]
    \centering
    \includegraphics[width=0.98\linewidth]{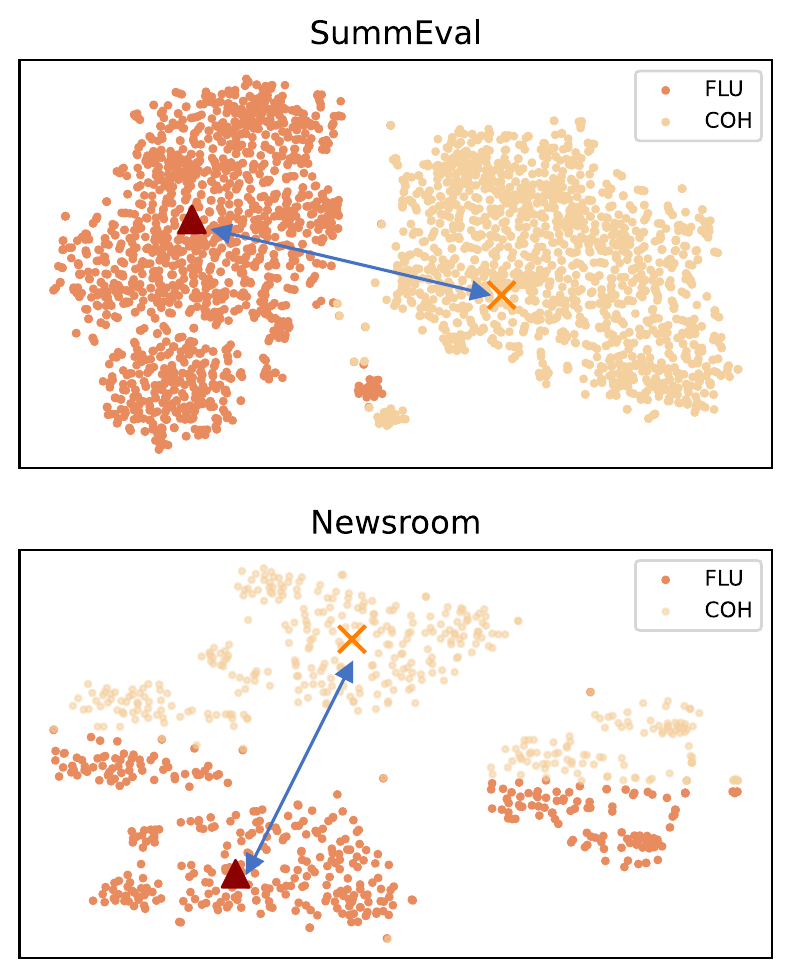}
    \caption{The t-SNE visualization of $rep$s shows the results of dimensionality reduction. The triangles and X on each figure represent the $rep$s of the same sample obtained using different prompts.}
    \label{fig:tsne_rep}
\end{figure}
\begin{figure*}
    \centering
    \includegraphics[width=0.99\linewidth]{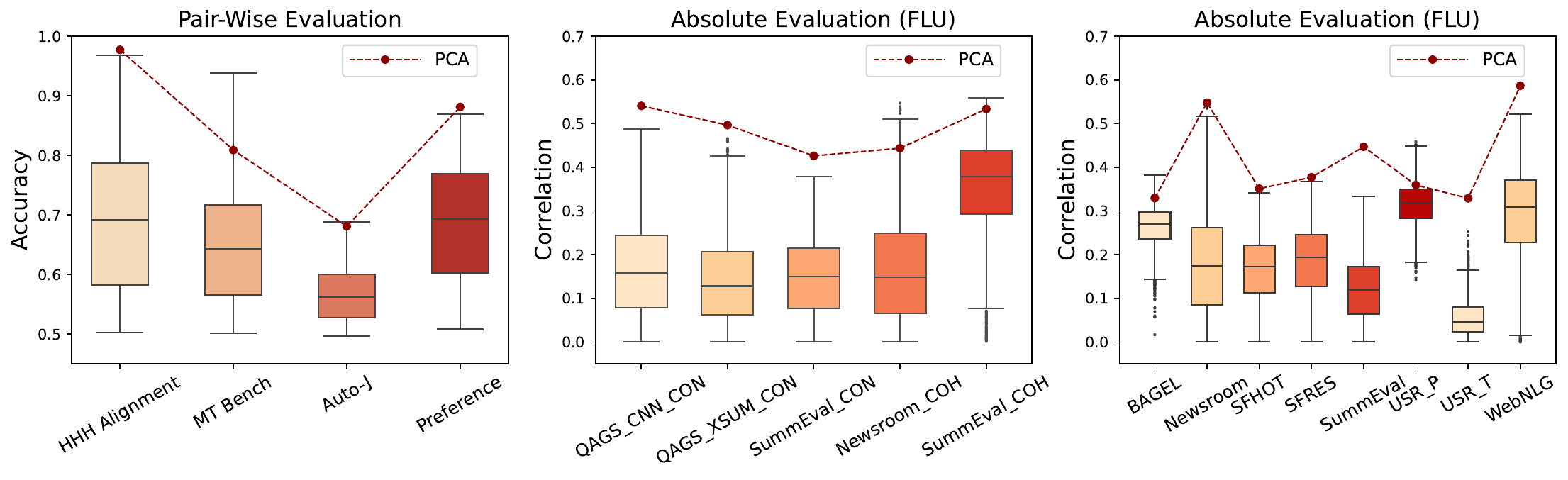}
    \caption{\textbf{Random Test Results} Box plots represent meta-evaluation results corresponding to random vectors $v$, while the scatter points in the figure represent the results corresponding to direction vector $d$ obtained through PCA. For pair-wise evaluation, the y-axis starts at 0.5, which is the expected accuracy of random guessing.}
    \label{fig:random_test}
\end{figure*}
\subsection{How prompt influence \textit{rep}s?}

The design of the prompt is an important step when applying RepEval for evaluation. Especially in absolute evaluation, when we need to evaluate different aspects of the same sample, we need to use different prompt templates to obtain the corresponding $rep$. However, what role do these prompts play? Do they truly distinguish between application scenarios? The previous experiments did not provide an answer to this question.

In this section, we utilize t-SNE for the visualization of $rep$s. We choose SummEval and Newsroom for this experiment, as they include evaluation results for two criteria: fluency and coherence. We collected $rep$s obtained from the two prompt templates and visualized their distribution using t-SNE, which is shown in Figure~\ref{fig:tsne_rep}.

It can be seen that representations collected from different prompt templates exhibit different distributions and can be clearly separated from each other. This indicates that the prompts successfully transfer $hyp$ to different positions within the semantic space, enabling the construction of the corresponding project direction in the transformed space and providing relevant assessments of the target criterion.

\subsection{Selection of Token and Layer}
To better utilize RepEval, in this section, we explore the performance of RepEval with different layers and token selections. Limited by space, we take fluency on absolute evaluation as an example and select four datasets from four tasks. All experiments follow the settings described in Section~\ref{sec:experiment}. The results are in Figure~\ref{fig:heatmap}.
 
The results show that, surprisingly, the last token is not always the best one. Moreover, the correlation scores increase sharply in the middle layers and achieve the best result. A possible explanation could be that $rep$s collected from middle layers contain more information relevant to the current context. Comparatively, $rep$s from the last layers are more useful to the next token prediction.

This provides us with the following suggestions for improving RepEval. Firstly, we can opt for the token in the last second or third position, instead of the last one token. Secondly, choose embeddings from the second half of the layers. The layer should be far enough from the input to ensure that sufficient information is encoded.

\subsection{A Good Projection or Not?}

Previous experiments show that PCA works effectively in identifying a suitable projection vector, surpassing other non-linear methods such as SVM. However, it remains uncertain whether PCA identifies the ``best'' projection. To address this question, we conduct the following experiments.

We randomly generated 2000 vectors $\Vec{v}$
with the same shape as the vector $\Vec{d}$ obtained by PCA in Section~\ref{sec:project_direction}. We then collect scores using the process outlined in Section~\ref{sec:collect_score}, replacing $\Vec{d}$ with $\Vec{v}$
The selection of token and layer positions followed the settings of PCA outlined in Section~\ref{sec:methodology}. The distribution of meta-evaluation results is shown in Figure~\ref{fig:random_test}.

We observe that $\Vec{d}$ obtained through PCA is a relatively optimal result. Compared to random vectors, it achieves nearly the highest correlation scores in absolute evaluation, as well as the highest accuracy scores in pair-wise evaluation. 
This indicates that if $rep$s contains related task information and that there exist projection vectors $\Vec{d}$ characterizing the direction of variation in text quality, PCA can efficiently help researchers find the target $\Vec{d}$, and be applied for evaluation.

\section{Conclusion}
We introduce RepEval, an evaluation metric utilizing the projection of LLM representations to obtain evaluation results, which exhibits a stronger correlation with human judgments in absolute evaluation, as well as higher accuracy in pair-wise selection than previous methods. RepEval is flexible and is easy to transfer to other evaluation scenarios, requiring only a few sample pairs for training, while avoiding the usage of LLMs with a large number of parameters such as GPT-4. We also provide suggestions on the proper application of RepEval, such as the selection of tokens and layers. Our work provides insights into the development of new metrics.

\section*{Limitations}
In this study, the language is restricted to English. Further research is necessary to validate the identified performance across a broader spectrum of tasks and languages.

The analysis in this study is primarily driven by experimental data, and we acknowledge the absence of a more comprehensive mathematical explanation of the underlying mechanisms of RepEval. Additionally, our evaluation relies solely on correlation and accuracy as measurement methods. A more detailed analysis is left for future work.

\section*{Acknowlegement}
This work was supported by  NSF China (No. 61960206002,62020106005, 42050105, 62061146002),   Shanghai Pilot Program for Basic Research - Shanghai Jiao Tong University
\bibliography{reference}

\clearpage

\appendix

\section{Experiment Settings}
When evaluating fluency and consistency, we construct the training dataset using Asset. For coherence evaluation, we utilize GCDC. During the training of the PCA model, the number of training pairs is set to 5 and 20. Additionally, we employ the SVM model for comparison with the PCA method, using 100 pairs for SVM training. As SVM needs more training data, during construction, we ensure the distinctiveness of each pair, though some pairs may contain the same good or bad text. No repeated data is contained in the training set of PCA.

We collected representations with Mistral-7b following the process described in Section~\ref{sec:collect_rep}. We employ the Sklearn implementation of PCA and SVM. For SVM, the kernel is set as Radial Basis Function (RBF), gamma = $1/d$, and the regularization parameter $C = 1$. We utilized Mistral-7b to generate representations using a single NVIDIA GeForce RTX 3090. The training of PCA and SVM models was performed on a CPU. More experiment details can be found in Appendix~\ref{app:exp}.

\section{Evaluation Criteria}

\paragraph{Coherence}
In accordance with \citet{dang2005overview}, coherence evaluates whether models generate a well-structured and organized text body that aligns with the given task, steering clear of a mere compilation of related information.

\paragraph{Consistency}
Consistency, as per \citet{honovich-etal-2022-true-evaluating}, assesses whether all factual information in the output text corresponds with the content provided in the input.

\paragraph{Fluency}
Fluency, as defined by \citet{kann2018sentence-fluency}, gauges the natural perception of a sentence by humans. In certain instances, fluency is also referred to as naturalness, grammaticality, or readability.

\section{Experiments}

\label{app:exp}
\subsection{Datasets}
\label{app:dataset}
\subsubsection{Absolute Evaluation}
\paragraph{ASSET}
ASSET is a dataset created for the tuning and evaluation of sentence simplification models~\cite{alva-manchego-etal-2020-asset}. 
In this research, we use the human rating corpus, which contains 100 pairs of original sentences and system simplification as well as the human evaluation results for the system output. For each pair, the rating is done by 15 crowd-sourced workers from 3 aspects: fluency, adequacy, and simplicity.

\paragraph{BAGEL}
BAGEL features annotations on data-to-text tasks gathered from a dialogue system, with human annotations covering informativeness and naturalness, according to \citet{mairesse-etal-2010-bagel}. In this context, informativeness is compared with the gold standard, differing from our defined usage. However, for our purposes, we solely utilize the judgment results related to naturalness.

\paragraph{GCDC}
GCDC is created with real-world texts, which is designed for the development of discourse coherence algorithms~\cite{lai-tetreault-2018-gcdc}. Each sample in GCDC contains three evaluation scores of coherence on a 3-point scale from 1 (low coherence) to 3 (high coherence).

\paragraph{NEWSROOM}
NEWSROOM gathers 60 articles along with summarization outcomes from 7 models, featuring human-written summaries as references, as documented by \citet{grusky2018newsroom}. The evaluation encompasses coherence, fluency, relevance, and informativeness.

\paragraph{QAGS}
QAGS encompasses reference texts and annotation results focused on consistency in the context of the summarization task, as outlined by \citet{wang-etal-2020-asking}. The approach involves collecting three annotations for each sentence in a generated summary, utilizing a majority vote strategy to determine a consistency score. The final score is obtained by calculating the mean value across all sentences.

\paragraph{SFHOT and SFRES}
SFHOT and SFRES deliver evaluation results for the data-to-text task, incorporating annotations of naturalness and informativeness, as detailed by \citet{wen-etal-2015-semantically}. In this context, informativeness gauges the consistent degree between sources and hypotheses. This dataset is utilized for analyzing consistency, while naturalness serves as a proxy for fluency.

\paragraph{SummEval}
SummEval offers a compilation of summarization outcomes produced by language models, as detailed by \citet{fabbri-etal-2021-summeval}. These models undergo training on the CNN/DailyMail datasets, as described by \citet{hermann2015teaching}, along with their corresponding reference texts. Each generated summary in the dataset includes score results from both expert annotators and crowd-workers, covering four dimensions: coherence, consistency, fluency, and informativeness.

\paragraph{USR}
The USR dataset offers evaluation results for the dialogue task across five aspects: fluency, coherence, engagingness, groundedness, and understandability. In alignment with the rephrasing strategy outlined by \citet{zhong-etal-2022-unieval}, the original aspects "maintains context" and "natural" are renamed as "coherence" and "fluency," respectively.

\paragraph{WebNLG}
WebNLG includes human evaluation results from the 2017 WebNLG Challenge, which focuses on the data-to-text task, as described by \citet{shimorina2019webnlg}. The candidate text undergoes evaluation based on three aspects: fluency, grammar, and semantics. In this context, fluency assesses whether a text is smooth and natural, and the fluency score is employed for experimentation purposes.

Features contained in each absolute evaluation dataset are listed in Table~\ref{tab:dataset_criteria}. With the exception of GCDC, all datasets include $src$.

\begin{table}[!h]
    \centering
    \small
    \tabcolsep 0.072in
    \begin{tabular}{lcccc}
    \toprule
                             & \textbf{COH}             & \textbf{CON}        & \textbf{FLU}       & \textbf{REF}        \\
    \midrule    
    
    \multicolumn{1}{l}{\textbf{summarization}} & &            &            &            \\
    
    -Newsroom                 & \checkmark      &            & \checkmark & \checkmark \\
    -QAGS                     &                 & \checkmark &            & \checkmark \\
    -SummEval                 & \checkmark      & \checkmark & \checkmark & \checkmark \\
    \midrule
    \textbf{data-to-text}       &                 &            &            &            \\
    
    -BAGEL                    &              &  & \checkmark & \checkmark \\
    -SFHOT                    &                 & \checkmark & \checkmark & \checkmark \\
    -SFRES                    &                 & \checkmark & \checkmark & \checkmark \\
    -WebNLG                   &                 &            & \checkmark & \checkmark \\
    
    \midrule
    \multicolumn{2}{l}{\textbf{dialogue}}  &            &            &            \\
    -USR-Persona                    & \checkmark      &            & \checkmark &  \checkmark  \\
    -USR-Topical                      & \checkmark      &            & \checkmark &  \checkmark  \\
    \midrule
    \multicolumn{1}{l}{\textbf{simplication}} & &            &            &            \\
    -Asset                &      &            & \checkmark &  \\
    \midrule
    \multicolumn{1}{l}{\textbf{other}} &&&&\\
    -GCDC                     & \checkmark      &            &  &    \\
    \bottomrule
  
  \end{tabular}
  \caption{Datasets and available features.}
  \label{tab:dataset_criteria}

\end{table}

\subsubsection{Pair-wise Evaluation}
\paragraph{HHH Alignment} 
HHH Alignment contains the evaluation result based on four criteria: helpfulness, harmlessness, honesty, and other, as well as the relevant 221 response pairs judged by human evaluators~\cite{askell2021hhhalign}.
\paragraph{MT Bench}
MT-bench consists of a series of open-ended questions that evaluate a chatbot’s multi-turn conversational and instruction-following ability, which collect 3,360 response pairs based on 80 prompts, as well as judgment from human evaluators~\cite{zheng2024mtbench}.
\paragraph{Auto-J}
A dataset constructed with massive real-world scenarios with human evaluation judgments, consisting of 58 prompts and 1,392 response pairs~\cite{li2023autoj}.
\paragraph{Preference Bench}
The preference bench contains 2000 response pairs, which are constructed based on 200 prompts and 200 evaluation criteria, as well as human judgments~\cite{kim2024prometheus}.

\subsubsection{ Resources}
The resources of all datasets we used are listed as follows.
\begin{itemize}
    \item Newsroom, SummEval, QAGS\_cnn, QAGS\_XSUM, SFHOT, SFRES are downloaded from source provided by~\citet{NEURIPS2021_bartscore}. The related URL is \url{https://github.com/neulab/BARTScore}.
    \item Asset and WebNLG is downloaded from source provided by~\citet{Scialom2021BEAMetricsAB}. The related URL is \url{https://github.com/ThomasScialom/BEAMetrics}. We delete empty reference sentences before applying.
    \item USR\_Topical and USR\_Persona are created by~\citet{mehri-eskenazi-2020-usr}. The related URL is \url{https://github.com/shikib/usr}.
    \item GCDC is created by~\citet{lai-tetreault-2018-gcdc}, and the URL is \url{https://github.com/aylai/GCDC-corpus}.
    \item HHH Alignment, MT Bench, Auto-J, and Preference Bench are downloaded from source provided by~\citet{kim2024prometheus}. The related URL is \url{https://github.com/prometheus-eval/prometheus-eval}.
\end{itemize}

\subsection{Implement of Baselines}
\label{app:metrics}
\begin{itemize}
    \item BARTScore is downloaded from \url{https://github.com/neulab/BARTScore}. We use the faithfulness-based variant based on "facebook/bart-large-cnn"\footnote{\url{https://huggingface.co/facebook/bart-large-cnn}} checkpoint~\cite{lewis2020bart}.
    \item BERTScore is downloaded from \url{https://github.com/Tiiiger/bert_score}. We use the F1 score calculated based on checkpoint "deberta-xlarge-mnli"\footnote{\url{https://huggingface.co/microsoft/deberta-xlarge-mnli}}~\cite{he2021deberta}.
    \item GPTScore is downloaded from \url{https://github.com/jinlanfu/GPTScore} and we use the checkpoint "gpt2-large"\footnote{\url{https://huggingface.co/gpt2-large}}~\cite{radford2019language}.
    \item UniEval is downloaded from \url{https://github.com/maszhongming/UniEval}. We use the "summarization" variant developed based on checkpoint "MingZhong/unieval-sum"\footnote{\url{https://huggingface.co/MingZhong/unieval-sum}}~\cite{zhong-etal-2022-unieval}.
    \item For metric BLEU and Meteor, we use the implementation provided by the python package NLTK~\cite{bird2009natural}. 
    
\end{itemize}

\subsection{SVM}
We also add experiments with the Support Vector Machine (SVM) for comparison. With representation $rep$ as inputs, the SVM method involves training a binary classifier on good-bad text pairs, and we use the probability of a text belonging to good text as the score result. To be specific, consider a specific text, denote the predicted probability of being good text as $p_1$, the predicted probability of being bad text as $p_0$, and the score satisfies
\begin{equation}
    score = p_1 / (p_0 + p_1) = p_1
\end{equation}

For each pair, we randomly select one from the good text and another from the bad text. 
\begin{table}[!h]
    \centering
    \begin{tabular}{cccc}
    \toprule 
    Dataset & Range & Low & High \\
    \midrule 
      Asset   & [1, 100] & 1 & 90\\
      GCDC   & [1,3] & 1 & 3 \\
      \bottomrule
    \end{tabular}
    \caption{Score range of dataset Asset and GCDC.}
    \label{tab:score_range}
\end{table}

\subsection{Selection of Token and Layer}
\label{sec:best_selection}
Here we present the optimal layer and token selections for different RepEval settings and the SVM method, where $k$ represents the number of components of PCA.

\begin{table}[h]
\small
\tabcolsep 0.072in
\begin{tabular}{c|cccccc}
\toprule
criterion            & model & pairs & prompt & k  & layer & token \\
\midrule
\multirow{4}{*}{FLU} & PCA   & 20    & yes    & 4  & -15   & -4    \\
                     & PCA   & 5     & yes    & 4  & -15   & -2    \\
                     & PCA   & 20    & no     & 3  & -21   & -1    \\
                     & SVM   & 100   & yes    & - & -2    & -2    \\
\midrule
\multirow{4}{*}{CON} & PCA   & 20    & yes    & 3  & -16   & -2    \\
                     & PCA   & 5     & yes    & 3  & -15   & -2    \\
                     & SVM   & 100   & yes    & - & -2    & -1    \\
\midrule
\multirow{4}{*}{COH} & PCA   & 20    & yes    & 4  & -9    & -2    \\
                     & PCA   & 5     & yes    & 2  & -1    & -2    \\
                     & PCA   & 20    & no     & 3  & -1    & -2    \\
                     & SVM   & 100   & yes    & - & -1    & -3   \\
                     \bottomrule
\end{tabular}
\caption{Selection of token and layer in absolute evaluation. Where k is the number of main components when using PCA.}
\end{table}

\begin{table}[h]
\small
\centering
\begin{tabular}{ccccc}
\toprule
model & pairs & k & layer & token \\
\midrule
PCA & 5 &  1 & -13 & -1 \\
PCA & 20 & 1 & -2 & -1 \\
\bottomrule
\end{tabular}
\caption{Selection of token and layer in pair-wise evaluation. Where k is the number of main components when using PCA.}
\end{table}
\subsection{Prompt Template of RepEval}
\label{sec:repeval_prompt}
As described in Section~\ref{sec:collect_rep}, the prompt templates of RepEval are listed as follows.
\subsubsection{Absolute Evaluation }
For all absolute evaluation, we use the same prompt template. 

\begin{myshaded}

\textit{Is the following Hyp <criterion\_description>? }

\textit{Hyp: <hyp>}

\color[RGB]{117, 113, 113} {
\textit{Src: <src> }
}

\color[RGB]{0, 0,0}
\textit{The sentence is}
\end{myshaded}
Apart from the inputs of $src$ in consistency evaluation, we only change the <criterion\_description> in the template, and please refer to Table~\ref{tab:criterion_description} for details.

\begin{table}[!h]
    \centering
    \begin{tabular}{cc}
    \toprule
    criterion & criterion description\\
    \midrule
      fluency   &  fluent \\
      coherence   & coherent \\
      consistency & consistent with Src \\
      \bottomrule
    \end{tabular}
    \caption{Criterion description for each criterion in absolute evaluation.}
    \label{tab:criterion_description}
\end{table}

\subsubsection{Pair-wise Evaluation}
Refer to the prompt design of ~\citet{kim2024prometheus}, we use the following prompt template for all pair-wise evaluation. Here, for pairs from different datasets, the score rubric should also be chanted to the related one.
\label{app:prompt_p}
\begin{myshaded}
\#\#\#Task Description:
An instruction (might include an Input inside it), a response to evaluate, and a scoring rubric representing evaluation criteria are given.

Choose a better response between Response A and Response B. You should refer to the scoring rubric.

\#\#\#Instruction:
You are a fair judge assistant assigned to deliver insightful feedback that compares individual performances, highlighting how each stands relative to others within the same cohort.

\#\#\#Response A:
{<hyp\_1>}

\#\#\#Response B:
{<hyp\_2>}

\#\#\#Score Rubric:
{<score\_rubric>}

\#\#\#Ans: """
\end{myshaded}

\subsection{Prompt of LLM-based Absolute Evaluation}
\label{app:prompt_a}
In this study, we use the gpt-3.5-turbo, gpt-4 API, and mistral-7b for a zero-shot baseline. Following the designs of~\citet{shen-etal-2023-llm-summ-eval}, the prompts we utilized for each criterion are listed as follows.

\subsubsection{Absolute Evaluation of Fluency}

\begin{myshaded}
Score the following sentence with respect to fluency with one to five stars, where one star means "disfluency" and five stars means "perfect fluency". Note that fluency measures the quality of individual sentences, whether are they well-written and grammatically correct. Consider the quality of individual sentences.

Summary: <hyp>

Stars:
\end{myshaded}

\subsubsection{Absolute Evaluation of Coherence}

\begin{myshaded}
Score the following text with respect to coherence with one to five stars, where one star means "incoherence" and five stars means "perfect coherence". Note that coherence measures the quality of all sentences collectively, to the fit together and sound naturally. Consider the quality of the sentences as a whole and just output an overall score and no more other.

Summary: <hyp>

Stars:
\end{myshaded}

\subsubsection{Absolute Evaluation of Consistency}

\begin{myshaded}
Score the following summarization given the corresponding article with respect to consistency with one to five stars, where one star means "inconsistency" and five stars means "perfect consistency". Note that consistency measures whether the facts in the summary are consistent with the facts in the original article. Consider whether the summary reproduces all facts accurately and does not make up untrue information.\\
Article: <src>\\
Summary: <hyp>\\
Stars:
\end{myshaded}

\end{document}